\documentclass{article}

\usepackage{arxiv}

\usepackage[utf8]{inputenc} 
\usepackage[T1]{fontenc}    
\usepackage{hyperref}       
\usepackage{url}            
\usepackage{booktabs}       
\usepackage{amsfonts}       
\usepackage{nicefrac}       
\usepackage{microtype}      
\usepackage{lipsum}
\usepackage{graphicx}
\graphicspath{ {./images/} }

\usepackage{amsmath}
\usepackage{amssymb}
\usepackage{tabularx}
\usepackage{array}
\usepackage[numbers]{natbib}

\usepackage{color,soul}

\title{Abstraction, Validation, and Generalization for Explainable Artificial Intelligence}

\author{
 Scott Cheng-Hsin Yang$^*$ \\
  Department of Mathematics and Computer Science\\
  Rutgers University\\
  \texttt{scott.cheng.hsin.yang@gmail.com} \\
  $^*$ Corresponding author
   \And
 Tomas Folke \\
  Department of Mathematics and Computer Science\\
  Rutgers University\\
  \texttt{tomas.folke@rutgers.edu} \\
  \And
 Patrick Shafto \\
  Department of Mathematics and Computer Science\\
  Rutgers University\\
  \texttt{patrick.shafto@rutgers.edu} \\
}

\date{September 2, 2021}

\begin{document}
\maketitle
\begin{abstract}
Neural network architectures are achieving superhuman performance on an expanding range of tasks. To effectively and safely deploy these systems, their decision-making must be understandable to a wide range of stakeholders. Methods to explain AI have been proposed to answer this challenge, but a lack of theory impedes the development of systematic abstractions which are necessary for cumulative knowledge gains. We propose Bayesian Teaching as a framework for unifying explainable AI (XAI) by integrating machine learning and human learning. Bayesian Teaching formalizes explanation as a communication act of an explainer to shift the beliefs of an explainee. This formalization decomposes a wide range of XAI methods into four components: (1) the target inference, (2) the explanation, (3) the explainee model, and (4) the explainer model. The abstraction afforded by Bayesian Teaching to decompose XAI methods elucidates the invariances among them. The decomposition of XAI systems enables modular validation, as each of the first three components listed can be tested semi-independently. This decomposition also promotes generalization through recombination of components from different XAI systems, which facilitates the generation of novel variants. These new variants need not be evaluated one by one provided that each component has been validated, leading to an exponential decrease in development time. Finally, by making the goal of explanation explicit, Bayesian Teaching helps developers to assess how suitable an XAI system is for its intended real-world use case. Thus, Bayesian Teaching provides a theoretical framework that encourages systematic, scientific investigation of XAI. 
\end{abstract}


\section{Introduction}
Over the past decade neural network architectures have had impressive performance gains, reaching human or even super-human performance in many tasks, including speech recognition, translation, and image classification \cite{sejnowski2020unreasonable}.
These architectures have the potential to revolutionize many human activities, including logistics, medicine, and law \cite{armour2020unlocking, amisha2019overview, winkelhaus2020logistics}; however, the responsible and safe deployment of these systems depend on them being understandable to human stakeholders \cite{atkinson2020explanation}.
Two solutions have been suggested to this problem: one is to design systems that are inherently interpretable or transparent, which often involves a compromise in performance; the second is to develop bespoke solutions to explain the decision-making of an obscure system post-hoc \cite{dovsilovic2018explainable}.
In this paper, we present a third approach in which explainability is analyzed as a problem of facilitating understanding of AI systems by humans. Thus, we propose a general approach to explaining AI systems by explicitly analyzing the problem of providing information that enables a human to understand and predict the AI. The aim of this article is to introduce a unifying framework to think about XAI in terms of decomposable components from a cognitive science perspective. We show how this framework provides novel insights to prior research by explicating and modularizing the different components of existing XAI systems. Once identified, these components can be validated, and we discuss the implications of such validation for the generalizability of XAI solutions, adding a new dimension to XAI research.

The literature on explainable AI has exploded in recent years \cite{arrieta2020explainable}, but there is still a dearth of coherent theoretical frameworks of XAI techniques \cite{gerlings2020reviewing}, and the taxonomies that do exist are based on the technological substrate underlying the explanation techniques as opposed to their pragmatic goals. 
This lack of theory hobbles XAI research because it obscures what lessons can be safely transferred between studies and applications, and which components need to be re-validated in new contexts.
As a consequence, it reduces both the speed of knowledge accumulation and deployment of safe, explainable AI systems across sectors. 
Additionally, most XAI solutions tend to be designed by software engineers for engineers, and as such do not consider how to explain the target system to non-technical users  \cite{miller2017explainable, miller2019explanation, abdul2018trends, poursabzi2018manipulating}.
This is problematic because successful explanation clearly depends on the users and their goals \cite{lipton2018mythos}, and if an AI system is successfully deployed, software engineers will be a small subset of the users. 

Explainable AI is a complex problem, with both technological and psychological components.
A theoretical framework that formulates the problem of XAI in structured and normative ways may surface associations between different methods and results that previously seemed disconnected.
Such a framework also decomposes XAI problems into abstractions that represent fundamental components and dependencies, that can be validated separately.
Furthermore, such a modular approach would support deployment, as it would allow formal testing as to what sub-components of explanation generalize to what contexts.
We propose Bayesian Teaching, which formalizes explanation as a social act between a teacher and a learner, as such a framework.
In the following section we will explain how Bayesian Teaching abstracts a wide range of XAI systems into the following four components (see Table~\ref{tab:glossary}): (1) the target inference, (2) the explanation, (3) the explainee model, and (4) the explainer model.
For concreteness, we show how to apply Bayesian Teaching to decompose popular classes of XAI methods.
Then, we illustrate how the decomposed parts can be validated semi-independently through user studies, and reflect on how Bayesian Teaching promotes human-centeredness in XAI research and application.
Finally, we discuss generalization of the decomposed parts, including remarks on the manipulation and recombination of the components.

\begin{table}[h!]
    \centering
    \begin{tabular}{ m{3cm} m{12cm}}
    Term & Definition\\
    \hline
    \hline
    Target inference, $\Theta$ & The inference about the AI system that the XAI system is designed to help a user reach. Examples: the classification decision, the decision rule, predictive distribution.\\
    \hline
    Explanation, $x$ & A stimuli presented to a human user with the aim to shift their beliefs towards the target inference.\\
    \hline
    Explainee/learner model, $P_L(\Theta|x)$ & Probabilistic formalization of the agent who updates their belief of the target inference based on the explanation received; in the XAI context, a model of the human user. \\
    \hline
    Explainer/teacher model, $P_T(x|\Theta)$ & Bayesian formalization of the agent that selects explanations for the learner to shift the learner's beliefs towards the target inference (see Equation~\ref{eq:BT}); in the XAI context, the XAI system.\\
    \hline
    \end{tabular}
    \caption{Glossary of abstracted XAI components in Bayesian Teaching.}
    \label{tab:glossary}
\end{table}

\section{Bayesian Teaching} \label{sec:bt}

Bayesian Teaching formalizes explanation as a communication act between the explainer (teacher) and the explainee (learner) by the following equation:
\begin{align}
P_T(x|\Theta) = \frac{P_L(\Theta|x)P(x)}{\sum_{x'\in\Omega}P_L(\Theta|x')P(x')}.
\label{eq:BT}
\end{align}
The equation describes how a teacher $P_T$ should select an explanation $x$ to best explain a target inference $\Theta$, contingent on their model of the learner $P_L$. Specifically, it says that the probability of choosing an explanation $x$ to explain the target inference $\Theta$ is proportional to the probability that the explanation $x$ would lead the learner model $P_L$ to the target inference $\Theta$. Thus, Bayesian Teaching explicitly decomposes the explanation generation process into four components: the target inference $\Theta$; the explanation $x$; the learner model, which is captured by the posterior $P_L(\Theta|x)$; and the teacher model, which is captured by the selection posterior $P_T(x|\Theta)$. The explanation is selected from a pre-specified set $\Omega$, in which each element has a prior probability $P(x)$ of being selected. Note that in our account the XAI system takes an active role in selecting explanations, while the human learner is assumed to be a passive recipient. Our framework can be extended to model settings where a human user interacts with an XAI system by actively querying it. The extended framework would involve formalizing the reduction of uncertainty about $\Theta$, which can be derived from the learner model $P_L(\Theta|x)$ \cite[e.g.,][]{crupi2018generalized}.

The aim of XAI is to improve human users' understanding of AI systems. As such, a successful explanation should shift the user's belief to increase the fidelity between their internal model of the AI system and the AI system itself. Bayesian Teaching provides a formal account of such intentional belief-shifting via explanation. One implication of the belief-shifting perspective is that the success of an explanation is determined by how much it shifts a user's belief in a desired direction. Bayesian Teaching specifies the components---$\Theta$, $x$, $P_L(\Theta|x)$, and $P_T(x|\Theta)$---required to quantitatively model this shift. Below we define and explain each of these components.

\textbf{The target inference} $\Theta$ is an aspect of the model that human users wish to understand. Possible target inferences range from global aspects of the model, such as model parameters, to intermediate components, such as the model's latent variables \cite{vong2018bayesian}, to local targets, such as the model's prediction on a particular data point \cite{yang2021mitigating}. Local targets can be curated into a curriculum to inform model behavior on a holistic level. The size and complexity of the curriculum capture the trade-off between explanation completeness and explanation complexity. In general, the target $\Theta$ is related to the behavior of the model to be explained. Specifically, one should consider how the target $\Theta$ aligns with the actual use cases in deployment, such as debugging, verification, or acceptance testing.

\textbf{The explanation} $x$ is the object provided to the end user to induce understanding about $\Theta$. Common explanation media include instances from the training data, features of the data (e.g. saliency maps), and simplified models that accurately describe the target model for some subset of the problem space. Toward the goal of enhancing understanding, a key consideration when picking an $x$ is ease of processing by human users. Appropriate media can often be derived from the model's training data, as they are typically curated by and hence understandable to humans. Using XAI of image classification as an example, one can take the training images at different level of granularity to produce different types of $x$, including the images themselves \cite{kim2016examples, yang2021mitigating}, regions of an image \cite{ribeiro2016should}, or pixels of an image \cite{lundberg2017unified}. Other media that are easy to process are intuitive decision rules, which can be captured as distilled rule sets or even entire models, such as decision trees and linear models as is done in mimic learning.

\textbf{The learner model} $P_L(\Theta|x)$ is a computational model that describes how the user makes inferences about $\Theta$ when given the explanations $x$. All XAI methods have a learner model, explicit or implicit. We advocate making the learner model explicit to allow validation of this crucial component. The quality of the explanation generated depends on the quality of the learner model used. An inaccurate learner model will lead to unreliable and confusing explanations because the mapping between $x$ and $\Theta$ would be inaccurate. Conversely, a perfect learner model could lead to optimal explanation. The specifications of $x$, $\Theta$, and $P_L(\Theta|x)$ provide the input, output, and the assertions to be tested in a validation. In much of the XAI literature the learner model is embodied as a loss function. The loss function composes of two parts: a \textit{mapping} from $x$ to a $\Theta'$ that exists in the same mathematical space as $\Theta$, and the loss part that can be used to assign a probability to the target $\Theta$ based on how different $\Theta'$ and $\Theta$ are \cite{murphy2012machine}.  

\textbf{The teacher model $P_T(x|\Theta)$} specifies the explanation-generation process. As Equation~\ref{eq:BT} suggests, this selection process is largely determined by the learner model. This is intuitive because a good teacher should consider the learner when selecting an explanation.
To find the optimal explanation, one could search for the $x$ that maximizes $P_L(\Theta|x)$, and hence $P_T(x|\Theta)$. This solution is equivalent to finding the $x$ that maximizes the numerator of Equation~\ref{eq:BT} and thus avoids the computation of the denominator. Other approaches to inference include sampling from the $P_T(x|\Theta)$, which will provide a sense of the relative effectiveness of near-optimal explanations.

\textbf{The $\Omega$ and $P(x)$ terms} describe additional constraints on the explanations considered in the explanation-generation process. $\Omega$ specifies the pool of explanations to select from, and $P(x)$ specifies which element in the pool is more likely \textit{a priori}. Factors that influence the design of $\Omega$ include cognitive load and intuition on regions on interest. Returning to the image classification example, one may want to show a limited number of images as explanations to avoid cognitive overload or limit saliency maps to only the lungs in an x-ray image for pulmonary disease detection. In the literature, $P(x)$ is often used to control the complexity of the explanation. For example, if $x$ are decision trees, shallower trees are given higher prior probability than deeper ones. $P(x)$ can also capture the cost for constructing the $x$ in terms of computational resources, manual labor, etc.

\section{Decomposition via abstraction}\label{sec:abstraction}
In the previous section we have identified abstractions central to Bayesian Teaching. Here we demonstrate how these abstractions facilitate decomposition of existing methods into component parts. We consider three popular classes of XAI methods: explanation-by-examples, explanation-by-features, and mimic learning. These three methods differ with regards to what explanatory medium they use, which is the most common distinguishing factor between current XAI methods. 
However, all also make commitments regarding the target inference $\Theta$, the learner model $P_L(\Theta|x)$, and the sampling of explanation in $P_T(x|\Theta)$, as we illustrate in the following subsections. 

\subsection{Explanation-by-examples}

Explanation-by-examples seeks to explain the behavior of a target model by presenting a subset of cases from the training data that strongly influenced the model's inference. For instance, to explain why a classifier categorized a certain image as a cat as opposed to other animals, examples of cat may be provided to show what the classifier considers as prototypical cats. The explanation-by-examples approach has many desirable properties: It is fully model-agnostic and applicable to all types of machine learning \cite{chen2018learning, eaves2016tractable, ho2016showing}; it is domain- and modality-general \cite{hendricks2018generating, kanehira2019learning}; and it can be used to generate both global explanation \cite{kim2014bayesian, kim2016examples, vong2018bayesian, wang2018dataset, koh2017understanding} and local explanation \cite{papernot2018deep, yeh2018representer, goyal2019counterfactual}. In the context of Bayesian Teaching, explanation-by-examples is obtained by setting $x$ to be a data point from the training data \cite{yang2017explainable}. Below we use the Bayesian Teaching framework to decompose two existing explanation-by-examples methods into the components described in Section~\ref{sec:bt}.

Suppose you have an image classifier and you would like to understand how the it represents its target classes. In a previous study based on Bayesian Teaching \cite{vong2018bayesian}, the authors address this issue by finding a few examples from the training data that captures the target model's class representations. Here the target $\Theta$ is the latent class means of a probabilistic linear discriminant analysis (PLDA) model trained on the whole training dataset. The explanation $x$ is a set of three images from the training dataset for each class. The learner model, $P_L(\Theta|x)$, computes the probability that a PLDA (the mapping) assigns to $\Theta$ when trained on $x$ instead of the full dataset. The teaching is based on max selection from $P_T(x|\Theta)$. The Bayesian Teaching paper \cite{vong2018bayesian} also proposes that the goodness of $x$ can be evaluated by whether the explanations help humans predict the target model's classification in a two-alternative-forced-choice (2AFC) task.

Similar to the previous method for explaining the class representation of a classifier, in the maximum-mean-discrepancy-critic (MMD-critic) method proposed in \cite{kim2016examples}, the authors aimed to find a small subset of data that can represent the entire data set (as opposed to the latent means of the classes). They did this by modelling the data distributions which allows for the comparison between the target distribution and the learner-model induced distribution needed for example selection. Here the model to be explained is a kernel function on top of a deep neural net (DNN), and the target $\Theta$ is the data distribution of a particular class induced by this model.
The explanation $x$ is a set of images from the training data, with $\Omega$ restricting the set size and the classes from which the images can be sampled. The learner model, $P_L(\Theta|x)$, is a function that describes how similar the data distribution induced by the kernel function (the mapping) on $x$ is to $\Theta$. The teaching process of selecting $x$ to induce the desired data distribution $\Theta$ is done via max selection. More specifically, the work presents two sub-targets: ``prototype" and ``criticism." The prototype selection aims to make the distribution induced by the learner model on $x$ match $\Theta$ as much as possible, whereas the criticism selection aims to make the two differ as much as possible. 
The MMD-critic paper \cite{kim2016examples} also presents an evaluation task that measures how much the explanations help humans predict the model's predictions in terms of accuracy and response speed.
To evaluate the model with human participants, the authors bridged the gap between $\Theta$ (a data distribution) and the evaluated class prediction by a nearest-neighbor-classifier based on the kernel distance in the DNN's final feature space.

\subsection{Explanation-by-features}

Explanation-by-features seeks to explain AI decisions by drawing attention to sub-components of an instance of the data that influence the output decision. In other words, this class of methods sets the $x$ to be a constituent of a data point, such as regions of an image, phrases in a document, or elements in a vector. Saliency maps are a form of explanation-by-features used in image classification, and is among the most popular XAI methods in recent years \cite{atrey2019exploratory}. Below we decompose two existing methods for generating saliency maps through the lens of Bayesian Teaching.

The Randomized Input Sampling for Explanation (RISE) method for generating saliency maps is a simple, model-agnostic method that only requires access to the output probabilities --- and not the internal workings --- of the target model to be explained \cite{petsiuk2018rise}. The target $\Theta$ is the predicted class label of a test image from the target model, typically a convolutional DNN. The explanation $x$ is a mask over the test image: pixels that are salient for the classification are fully unmasked. The learner model, $P_L(\Theta|x)$, outputs the predictive probability of the class label specified in $\Theta$ by passing the test image masked by $x$ into the target model (the mapping). Teaching selects a mask by finding the \textit{expectation} over the mask for a particular class. While finding a mask that maximizes $P_T(\Theta|x)$ is possible, that mask is likely to focus on a single salient region, whereas the expectation of the mask is likely to capture \textit{all} salient regions. Indeed, despite differences of algorithmic approach, RISE can be viewed as a special case of Bayesian Teaching \cite{yang2021mitigating}.

In the SHapley-Additive-exPlanation (SHAP) method presented in \cite{lundberg2017unified}, the authors generate feature saliency values, which are the weights of a linear model that locally matches the target model's inferences. Here, the target $\Theta$ is to perfectly match the target model's predictive distribution on a test data point.
The explanation $x$ is the saliency of features, such as individual pixels of an image or words in a document.
The learner model, $P_L(\Theta|x)$, assumes an additive linear model (the mapping), where the saliency values $x$ are the weights of the linear model. The learner model assumes two additional constraints referred to as missingness and consistency. Given the form of the model and the target $\Theta$ to perfectly match a distribution, the authors prove that there is only one solution (ie., one set of weights) that satisfy these constraints. This translates to the teaching process, $P_T(x|\Theta)$, being a delta function on the solution.
The SHAP paper \cite{lundberg2017unified} evaluates the goodness of $x$ by asking human participants to assign weights to certain features in order to compare the human-assigned weights to those generated by the method. Note that the SHAP method is closely related to the LIME method of \cite{ribeiro2016should}, which we will discuss in the following subsection. This connection provides a link between how explanation-by-features and mimic learning may morph into each other.

\subsection{Mimic learning}

Mimic learning, sometimes referred to as model distillation \cite{xie2020explainable}, is a class of explanation methods where the behavior of a complex, obscure model is approximated by a simpler model that is easier for humans to interpret \cite{adadi2018peeking}. Popular examples are approximating the local behavior of a deep neural network with linear models \cite{ribeiro2016should} or decision-trees \cite{frosst2017distilling}. In terms of the components introduced in Section~\ref{sec:bt}, the explanation $x$ is usually also the parameters of the learner model $P_L(\Theta|x)$ in mimic learning.

The pioneering LIME method \cite{ribeiro2016should} uses linear models to approximate the behavior of the target model locally. Here, the target $\Theta$ is the decision boundary of the target model in the neighborhood of a test data point, where neighborhood is defined by a combination of the kernel function and data augmentation used. The explanation $x$ is the weights of a linear model. This linear model maps the weights $x$ to a linear decision boundary. The learner model, $P_L(\Theta|x)$, is a function that quantifies how well the mapped decision boundary matches the decision boundary of the target model.
The teaching component of weight selection is done via maximizing $P_T(x|\Theta)$. To transform the weights into a form that can be easily processed by end users, the authors presented the features---such as regions of an image or words in a document---that have positive weights. \footnote{Strictly speaking, this visualization invokes another Bayesian Teaching problem where the $x$ are the features, and the rest of the components are left to be specified.} The LIME paper \cite{ribeiro2016should} presents three high-level tasks to evaluate explanations: participants were tasked with competency testing (predicting which classifier generalizes better), feature debugging (identifying harmful features), and model anomaly detection (identifying classifier irregularities).

The work in \cite{frosst2017distilling} distills a neural network into a soft decision tree. The target inference $\Theta$ is the predictive distribution given by the target DNN on a set of test data points. The explanation $x$ is a soft binary decision tree, which needs to be further visualized when provided to a human end-user. The learner model, $P_L(\Theta|x)$, is a function that quantifies the match between the prediction distribution from the soft binary decision tree with parameter $x$ (the mapping) and that from the target DNN to be explained. $P(x)$ is set to favor trees that have high-entropy in the paths taken over the target test dataset. The teaching process of selection $x$ is done via maximizing $P_T(\Theta|x)$.

\subsection{Summary}

In this section we have demonstrated how existing XAI methods can be broken down into components according to the abstraction provided by Bayesian Teaching. Such demonstration highlights that all XAI methods necessarily make commitments about $\Theta$, $x$, $P_L(\Theta|x)$, and $P_T(x|\Theta)$, and that these commitments can be made explicit. Having done the decomposition, one can now validate and ground each component, as discussed in the next section.
The decomposition resulting from the abstraction provides a template to reason about which insinuations of the different components can fit together.
Component-specific validation tests the extent to which a component would generalize across selections of other components and use cases.
Therefore, the abstraction, validation, and generalization made possible by Bayesian Teaching underscore the usefulness of our framework in systematizing research and development.

\section{Validation of components}\label{sec:validation}

In software development a \textit{unit test} is a piece of code that invokes one unit of work and checks whether that unit operates as intended.
When developing a new function it is considered good practice to write a unit test prior to writing the function itself \cite{olan2003unit}.
One reason for starting with designing the unit test is that it assures the programmer that they are trying to solve a well-specified problem and that they will know when they have succeeded.
Unit-testing in XAI is desirable for similar reasons.
The strict control a unit-test offers in software development might be unrealistic for much of XAI work, but as an analogy it captures what we should aim for.
Specifically, we want to systematically test the dependency between inputs and outputs of a specific XAI component.
This is different from a software unit test, in that components cannot always be individually evaluated the way software units can, but a test can still be designed to evaluate one specific component. Just as unit-tests allow for robust integration of functions, appropriate evaluation of XAI components supports their generalization beyond the original contexts in which they are introduced. 

In software development, a successful unit-test determines that the evaluated unit produces the desired input-output dependency and nothing else.
As we have alluded to throughout this article, the aim of XAI is to move the human user's beliefs so that they accurately capture the AI system; hence, any unit test of an XAI component must be a user study.
The specific design of the user study and the metrics recorded should be informed by the real-world use case the XAI system aims to solve: such as tweaking an AI system prior to release, improving human-AI teaming, or enabling effective auditing.
When AI is to influence high-stakes decisions in areas such as law and medicine, it might be helpful to specify an assertion statement that puts thresholds on the metrics evaluated.
For example, users need to accurately catch the AI's mistakes 95\% of the time, or they must have a normalized ranking loss below .05 when ranking the importance of the input features on a given decision.
The fact that one component passes a unit-test does not imply that it is suitable to explain any XAI model, for any user, with any goal, just like a software function passing a unit-test does not guarantee it will work in every software package.
However, a successful unit-test provides structured evidence that one component of the XAI solution meets some clearly defined criterion, and specifies the input-output interface that ensures the proper working of that component.
This information enables researchers and engineers to effectively reason about whether a component is appropriate for their application, and if it is, they can confidently integrate it into their application, leading to significant reductions in development time.

\subsection{Validation of Target Inference}

Evaluation of the target inference $\Theta$ requires $x$ because $x$ is the communication medium between the user and XAI system. However, in this case $x$ is auxiliary in that it does not \textit{need} to come from a sensible learner model, hence making the test independent of the learner-model component. The assumption of independence of the learner model is reasonable when the range of $x$ involved in the task can be efficiently covered by uniform sampling. Such evaluation tests whether it is possible to achieve $\Theta$ by \textit{some} $x$, but does not speak to the optimality of a \textit{specific} $x$.

To evaluate the modularity of $\Theta$, we can run a user study to measure a particular user metric on a given set of $\Theta$'s and $x$'s.
For example, one can design the set of $\Theta$'s to be the predicted classes of three test images from three different classes, the set of $x$'s to be example sets with different set size (e.g., 2, 3, or 4 images in a set), and the user-study metric to be the probability that the user would predict a $\Theta$ given an $x$.
If all the $x$'s result in the same rank-order of $\Theta$'s on the performance metric (e.g., if human's prediction is always poor on a particular class regardless of the number of examples shown), we can conclude that that $\Theta$ is independent of the set of $x$'s considered.
On the other hand, if the rank-order of $\Theta$ varies by $x$, the two should always be considered as a unit. Generalizing further, if the independence of $\Theta$ is established over a wide range of $x$, it increases the probability that it will work similarly with a novel, yet untested $x$; otherwise, each novel pairing need to be explicitly evaluated.

In evaluating $\Theta$, it is also important to distinguish between two forms of belief-updating in the context of XAI application. When auditing AI systems, the assumption is that the human user has access to the ground truth. Consequently, the XAI solution should help the user to develop an accurate model of the AI's ``beliefs," captured by the user's ability to predict the AI's generalization behavior. In this case, $\Theta$ would be focused on predictive behavior, as in most of Table~\ref{tab:components}.
When collaborating with AI systems, humans may have worse access to the ground truth than the AI. Consequently, the XAI and AI systems should work together to improve the human perception of the ground truth. In this case, in addition to understanding the AI, $\Theta$ should also include how the explanations lead human users to the ground truth. Providing explanation for teaching the ground truth returns Bayesian Teaching to its root in pedagogical modelling and shows how XAI and pedagogy are closely related.

\subsection{Validation of Explanation}

The medium of explanation can be evaluated partly independently from all other components of Bayesian Teaching. The reason for this partial independence is that explanation media can be evaluated along a dimension that is task-independent: ease-of-processing. It is self-evident that all else being equal, an explanation that is easy for humans to process should be preferred. Ease-of-processing can be decomposed into two primary elements: alignment with the human cognitive-perceptual system, and complexity. Different representations of the same underlying information can vary in human ease-of-processing. For example, humans can interpret probabilistic information more accurately when it is presented as natural frequencies rather than conditional probabilities \cite{gigerenzer2003simple}, especially if the natural frequencies are visualized as icon arrays \cite{garcia2010icon, galesic2009using}. Research on data-visualisation and risk-perception has studied how to optimize information presentation for human understanding \cite{maceachren2005visualizing, brodlie2012review, spiegelhalter2011visualizing}, and XAI researchers could benefit from implementing these lessons.

Whereas information should always be presented in a way that is maximally interpretable to humans, information complexity involves a trade-off. More-complex information comes at a processing cost, but complexity can also add value. The solution to this trade-off is constrained by the complexity of the target inference to be explained $\Theta$ and the processing capacity of the target users. Domain experts tend to have a greater tolerance for complexity in their area of expertise, possibly because they have developed strategies to chunk task-relevant information more efficiently \cite{chase1973perception, gobet1998expert}. To summarize, the appropriate complexity of an explanation is determined by the complexity of $\Theta$, as well as the capacity of the users, as such complexity need to be evaluated with reference to a specific $\Theta$, on the target population for the intended XAI solution. 

There have been some attempts to create unit-tests for explanation media without reference to a specific learner model \cite{lage2019evaluation, poursabzi2018manipulating}. For example, Lage and colleagues evaluated explanation media that varied on multiple complexity metrics. Their evaluation consisted of three different tasks: prediction (predicting the AI system's decision), verification (determining whether the system made an accurate decision), and counter-factual analysis (determining whether changing a single input feature would alter the system's classification). 
They measured the response time, accuracy, and subjective satisfaction of human participants in each condition. This provides a strong template for how to evaluate explanation media, but their results are not easily interpretable because they did not account for the complexity trade-off discussed previously.

\subsection{Validation of Learner Model}

The accuracy of the learner model $P_L(\Theta|x)$ is essential for effective explanation in the Bayesian Teaching framework. 
In order to be effective the learner model should capture both the user's belief prior to exposure to explanation, as well as the inferential processes the user applies to update their beliefs given the explanation. Thus, both the prior belief $P_L(\Theta)$ (implicit in a Bayesian learner model) the posterior beliefs $P_L(\Theta|x)$ should be evaluated. Because of this focus on human belief, both the development and the optimization of learner models depend on lessons from cognitive science as well as computer science. How well the learner model aligns with actual users can be evaluated by the \textit{fidelity} between the modelled response to a given explanation $x$ and the user's actual response. This involves specifying a $\Theta$, sampling $x$, computing $P_L(\Theta|x)$, and running a user study that measures the fidelity between the learner model and the actual intended users. When assessing the calibration of the learner model, it is important to cover a wide range of $P_L(\Theta|x)$ \cite{yang2021mitigating}. This implies evaluating both explanations that the learner model predict will improve user understanding and explanations expected to be detrimental. If the learner model accurately predicts the full range of user behavior in response to explanation, it can be considered to have passed the key test.

Different users vary in their prior beliefs, inferential biases, and goals. As such, it is often unrealistic to develop one general learner model that captures all users well. Instead, we propose to develop different learner models for different user classes, such as AI engineers and clinicians. The modular nature of the Bayesian Teaching framework often allows for varying the learner model $P_L(\Theta|x)$ to fit the current user, while keeping $\Theta$ and $x$ constant.

Aside from providing targeted explanations, formal learner models can add value to explainable AI by encoding general human inferential biases.
For example, recent evidence from human reinforcement learning suggests that people learn more from information that supports their existing beliefs, relative to information that contradicts them \cite{palminteri2017confirmation, tarantola2021confirmation}. A well-designed learner model should incorporate such biases, so that they can be accounted for and leveraged for effective explanation. We recommend a modular encoding of such biases that can easily interface with any learner model, treating it as a meta-model, so as to speed up the development cycle of new XAI solutions.

\subsection{Validation of Teacher Model}

The evaluation of the teaching process, $P_T(x|\Theta)$, by definition depends on the specified $x$, $\Theta$, and probability $P_T$. Equation~\ref{eq:BT} shows that teaching is fully determined by $\Theta$, $x$, the learner model given by $P_L(\Theta|x)$, $\Omega$, and $P(x)$. Thus, validation of these components implies the validation of the teaching process. By virtue of Bayes' rule, a user task suitable for evaluating the learner model will also be suitable for evaluating the teaching process. The former should focus on covering a wide range of $P_L(\Theta|x)$, while the latter should ensure the achievement of a particular $P_L(\Theta|x)$. 

Additionally, Bayesian Teaching defines the effectiveness of an explanation by the extent to which it shifts a user's belief towards a target inference. The belief-shifting framework puts an upper bound on explanation effectiveness: If the user holds the target belief prior to being exposed to the explanation, there is no way to measure a potential positive impact of the explanation. As such, it only makes sense to test XAI interventions when there is a misalignment between user beliefs about the target AI system and the ground truth.

\subsection{Validation on Use Cases}
The appropriate target inference $\Theta$ is determined by the use case the user wishes to solve. To give an example, suppose a doctor wishes to understand why an AI image classifier diagnosed a particular patient with cancer, the XAI system may want to expose the decision-boundary around that particular image. In this case, a target inference that focuses on the decision boundary would be more suitable than one that focuses on matching the modeled data distribution.

There can be multiple ways to design the $\Theta$ and $x$ to address the same use case. The pros and cons of each design often need to be evaluated empirically with user studies. Returning to the previous example of understanding the decision boundary around a data point, one can also set up a Bayesian Teaching problem to show what variation of masks would change the class label on the test data point from one to the other \cite[c.f.,][]{goyal2019counterfactual}. This design would correspond to setting $\Theta$ to be the different class labels (e.g., positive and negative) and $x$ to be a set of masks. Yet another way to explain the decision boundary is to give example data points from both side of the decision boundary. In this case, $\Theta$ could be the predicted label of the test data point, $x$ would be a set of data-label pairs, and $\Omega$ or $P(x)$ can be used to enforce data-label pairs from both classes \cite{folke2021ptx}.  

Use cases are often arranged in a hierarchy, where the higher-level use case depends on the performance in the lower-level use cases. This hierarchy can lead to a hierarchy of nested Bayesian Teaching problems. Returning to our medical example, a chief radiologist at the same hospital might want to prioritize what images should be passed along to a radiologist and what images can be automatically classified by the AI system. Here, the target inference relate to the general decision-boundaries of the model (not a specific image), with a special focus on determining regions where the decision-boundaries are poorly aligned with the ground truth (cases when the AI system are likely to make mistakes). One may tackle this use case by designing a two-level Bayesian Teaching problem, where the lower level problem aims to explain decision boundary around a data point as described above, and the higher level problem aims to curate the appropriate data points to cover the overall boundary. In the higher level problem, $\Theta$ can thus be the overall decision boundaries, $x$ is a set of local decision boundaries, and the learner model describes how the local decision boundaries are stitched together.

\section{Generalization}\label{sec:generalization}
In Section~\ref{sec:bt} we showed that Bayesian Teaching provides an abstract template to think about XAI by highlighting that explanation is a kind of goal-directed communication between the XAI system and human.
In Section~\ref{sec:abstraction} we showed how to break down existing XAI methods into components according to this abstract template. From the decomposition, two complementary threads emerge. The first is the validation of components. In Section~\ref{sec:validation} we argued that validation should be done and that \textit{modular} testing of certain components is possible. The second thread is the generalization of the components by means of recombining them to form novel XAI methods, which we discuss below. The two threads are complementary in that validated components would promote the validity of the recombination, \textit{even absent a holistic testing of the system}. However, because evaluation of individual components is rarely done, below we simply entertain the recombination and leave their validity as a research direction.

Table~\ref{tab:components} shows the components from the decomposition via abstraction in Section~\ref{sec:abstraction}. Using Table~\ref{tab:components}, we make some remarks about trends that surfaced from the decomposition, including novel methods that could be formed by recombination. 

\begin{table}[h!]
\begin{center}
\begin{tabular}{  m{0.3cm}  m{4.8cm}  m{2.5cm}  m{2.2cm}  m{4cm}  } 
   & Target inference $\Theta$ & Explanation $x$ & Learner model $P_L(\Theta|x)$ & Human evaluation task\\ 
 \hline\hline
 (i) & Latent variables of the target model (e.g., class means of PLDA) \cite{vong2018bayesian} & Images from training data \cite{vong2018bayesian,kim2016examples} & PLDA \cite{vong2018bayesian} & Predicting the target model's predictions \cite{vong2018bayesian,kim2016examples} \\ 
 \hline
 (ii) & Modeled data distribution of a particular class \cite{kim2016examples} & Mask over target image \cite{petsiuk2018rise, lundberg2017unified} & Kernel function on DNN \cite{kim2016examples} & User-assigned salience \cite{lundberg2017unified} \\ 
 \hline
 (iii) & Predicted class labels on target data points \cite{petsiuk2018rise} & Model parameters \cite{ribeiro2016should, frosst2017distilling}& DNN classifier \cite{petsiuk2018rise} & Competency testing: which classifier generalizes better \cite{ribeiro2016should} \\ 
 \hline
 (iv) & Predictive distributions given by the target model on target data points \cite{lundberg2017unified,frosst2017distilling} & Features (e.g., regions of an image) \cite{ribeiro2016should} &  Additive linear model \cite{lundberg2017unified,ribeiro2016should} & Feature debugging: identify harmful features \cite{ribeiro2016should} \\
 \hline
 (v) & The decision boundary of the target model in the neighborhood of target data points \cite{ribeiro2016should} &  & Soft decision tree \cite{frosst2017distilling} & Model anomaly detection: identify classifier irregularities \cite{ribeiro2016should} \\ 
 \hline
\end{tabular}
\end{center}
\caption{\label{tab:components}A list of the components obtained from existing XAI methods presented in \cite{vong2018bayesian,kim2016examples,petsiuk2018rise,lundberg2017unified,ribeiro2016should,frosst2017distilling}. For brevity, the mapping of the learner model that maps $x$ to $\Theta$ is listed, but the probability assignment is left out. In addition to the target inference $\Theta$, explanaiton $x$, and learner model $P_L(\Theta|x)$ mentioned in Section~\ref{sec:bt}, the human evaluation tasks used in these papers are listed. The teacher's selection $P_T(x|\Theta)$ is left out, because most of the methods investigated maximize that probability.}
\end{table}

\textbf{Explanation.} Most XAI methods are distinguished or classified by the $x$ component. We argue that a coherent theory that includes all necessary components is more desirable. The coherence and completeness of such a perspective not only clarifies the connections among XAI methods but also offer systematic guidance on how to manipulate and compose them effectively. 

\textbf{Teacher component.} Most methods select explanations by maximizing $P_T(x|\Theta)$. In the likely case of uncertainty in learner model and complexity penalty, strictly maximizing can lead to consistent suboptimal explanation. In general, max selection is not robust when the modeled inference of the user does not fully match that of the actual user \cite{wang2020mathematical}. Probabilistic selection, though less convenient, can lead to better average performance in these cases, and naturally supports testing whether explanations that are predicted to be effective indeed are. 

\textbf{$\Theta$ and evaluation task.} Not all $\Theta$ and evaluation task are equally well aligned. For example, the goal of explaining predicted class labels in $\Theta$ (iii) is very well aligned to the user task of predicting the target model's predictions in evaluation task (i). In contrast, the predictive label target in $\Theta$ (iii) and the selection of a competent model in evaluation task (iii) are not directly related. Such misalignment requires additional consideration: is it possible that a simple modification of $\Theta$ could improve this alignment? For example, would it be better to modify $\Theta$ to be the \textit{difference} between the two target models' predictive probability on some test data points? An alternative solution would be to select a curriculum of target data points that will help the user to figure out the more competent model. Such curricula can be fleshed out in another Bayesian Teaching problem, where the new $x$ is a sequence of $\Theta$, and the new $\Theta$ will explicitly specify measures of discrimination between the two models.

\textbf{Recombination.} If a $\Theta$ is on the level of the generalization behavior of the model in data space [e.g., $\Theta$, in Table~\ref{tab:components} (iii)--(v)], this $\Theta$ will work together with any $x$. This observation allows for the generation of new XAI methods from recombination. For example, one can find a soft decision tree [$x$, in Table~\ref{tab:components} (iii)] to best match the decision boundary of a target model [$\Theta$, in Table~\ref{tab:components} (v)]. On the other hand, if $\Theta$ is on the level of a model's latent variable or parameters [e.g., $\Theta$, in Table~\ref{tab:components} (i)], both the $x$ and the mapping of the learner model are limited to models with the same parametric form. In summary, this illustration of how Bayesian Teaching can decompose and recombine existing XAI methods is beginning to hint at how Bayesian Teaching can identify reusable, modular components akin to a software development process.

\section{Conclusion}
We have argued for the importance of the development of systematic abstractions, validation, and generalization for explainable AI. Whereas the current state of the art in the field depends on solutions that are problem-specific, rendering the study of XAI to be a series of unconnected engineering problems, we advocate for a systematic, scientific approach to XAI. We present a theoretical framework based on Bayesian Teaching, which unifies the human and machine aspects of the problem and is strongly supported by research in cognitive science. Bayesian Teaching introduces a collection of abstractions that facilitate systematic thinking about XAI, which allows for the decomposition of diverse prior approaches. The abstractions presented by Bayesian Teaching support systematic validation of components, a necessary aspect of modern software development, which ensures that component parts of our model behave as expected through component-wise user studies. We further argue that abstraction and validation together support generalization---the ability to recompose validated aspects of models into new XAI methods for rapid deployment on new tasks and domains. Through this systematic approach, Bayesian Teaching supports the a cumulative science of XAI that incorporates best practices of behavioral research and software development. 

\section*{Acknowledgements}
This material is based on research sponsored by the Air Force Research Laboratory and DARPA under agreement number FA8750-17-2-0146 to P.S. and S.C.-H.Y. The U.S. Government is authorized to reproduce and distribute reprints for Governmental purposes notwithstanding any copyright notation thereon.

This work was also supported by DoD grant 72531RTREP, NSF MRI 1828528 to P.S. The methods described herein are covered under Provisional Application No. 62/774,976. 

\section*{Conflict of Interest}
The authors declare no conflict of interests.

\bibliographystyle{unsrt}  
\bibliography{main}

\end{document}